\documentclass[journal]{IEEEtran}
\usepackage{graphicx}
\usepackage{amsmath}
\usepackage{amssymb}
\usepackage{makeidx}
\usepackage{subfig}
\usepackage{float}
\usepackage{tabularx}
\usepackage{makecell}
\usepackage{multirow}
\usepackage{algorithm}
\usepackage{algpseudocode}
\usepackage{hyperref}
\usepackage[justification=centering]{caption}
\usepackage{nth}
\usepackage{balance}

\begin{document}
\title{Green Stability Assumption: Unsupervised Learning for Statistics-Based Illumination Estimation}
\author{{%
Nikola Bani\'{c} and Sven Lon\v{c}ari\'{c}}%
}

\maketitle

\renewcommand\footnoterule{{\hrule height 0pt}}

\renewcommand\footnoterule{{\hrule height 0pt}}
\let\thefootnote\relax\footnotetext{

This work has been supported by the Croatian Science Foundation under Project IP-06-2016-2092.

Copyright (c) 2017 IEEE. Personal use of this material is permitted. However, permission to use this material for any other purposes must be obtained from the IEEE by sending a request to pubs-permissions@ieee.org.

The authors are with the Image Processing Group, Department of Electronic Systems and Information Processing, Faculty of Electrical Engineering and Computing, University of Zagreb, 10000 Zagreb, Croatia (email: nikola.banic@fer.hr; sven.loncaric@fer.hr).
}

\begin{abstract}
In the image processing pipeline of almost every digital camera there is a part dedicated to computational color constancy i.e. to removing the influence of illumination on the colors of the image scene. Some of the best known illumination estimation methods are the so called statistics-based methods. They are less accurate than the learning-based illumination estimation methods, but they are faster and simpler to implement in embedded systems, which is one of the reasons for their widespread usage. Although in the relevant literature it often appears as if they require no training, this is not true because they have parameter values that need to be fine-tuned in order to be more accurate. In this paper it is first shown that the accuracy of statistics-based methods reported in most papers was not obtained by means of the necessary cross-validation, but by using the whole benchmark datasets for both training and testing. After that the corrected results are given for the best known benchmark datasets. Finally, the so called green stability assumption is proposed that can be used to fine-tune the values of the parameters of the statistics-based methods by using only non-calibrated images without known ground-truth illumination. The obtained accuracy is practically the same as when using calibrated training images, but the whole process is much faster. The experimental results are presented and discussed. The source code is available at \url{http://www.fer.unizg.hr/ipg/resources/color_constancy/}.
\end{abstract}


\IEEEpeerreviewmaketitle

\section{Introduction}
\label{sec:introduction}

\IEEEPARstart{R}{egardless} of the influence of the scene illumination, the human visual system can recognize object colors through its ability known as color constancy~\cite{ebner2007color}. In the image processing pipeline of almost every digital camera there is also a part dedicated to computational color constancy~\cite{kim2012new}. It first estimates the scene illumination and then uses it to chromatically adapt the image i.e. to correct the colors. For a more formal problem statement an often used image $\mathbf{f}$ formation model written under Lambertian assumption is~\cite{gijsenij2011computational}
\begin{equation}
\label{eq:image}
f_c(\mathbf{x})=\int^{}_{\omega} I(\lambda, \mathbf{x}) R(\lambda, \mathbf{x}) \rho_c (\lambda) d\lambda
\end{equation}
where $c\in\{R, G, B\}$ is a color channel, $\mathbf{x}$ is a given image pixel, $\lambda$ is the wavelength of the light, $\omega$ is the visible spectrum, $I(\lambda, \mathbf{x})$ is the spectral distribution of the light source, $R(\lambda, \mathbf{x})$ is the surface reflectance, and $\rho_c(\lambda)$ is the camera sensitivity of color channel $c$. Assuming uniform illumination for the sake of simplicity makes it possible to remove $\mathbf{x}$ from $I(\lambda, \mathbf{x})$ and then the observed light source color is given as
\begin{equation}
\label{eq:e}
\mathbf{e}=\left(\begin{array}{c}e_R\\e_G\\e_B\end{array}\right)=\int^{}_{\omega} I(\lambda)\boldsymbol{\rho}(\lambda)d\lambda.
\end{equation}

The direction of $\mathbf{e}$ provides enough information for successful chromatic adaptation~\cite{barnard2002comparison}. Still, calculating $\mathbf{e}$ is an ill-posed problem because only image pixel values $\mathbf{f}$ are given, while both $I(\lambda)$ and $\boldsymbol{\rho}(\lambda)$ are unknown. The solution to this problem is to make additional assumptions. Different assumptions have given rise to numerous illumiantion estimation methods that can be divided into two main groups. First of these groups contains low-level statistics-based methods such as White-patch~\cite{land1977retinex, funt2010rehabilitation} and its improvements~\cite{banic2013using, banic2014color, banic2014improving}, Gray-world~\cite{buchsbaum1980spatial}, Shades-of-Gray~\cite{finlayson2004shades}, Grey-Edge~(\nth{1} and \nth{2} order)~\cite{van2007edge}, Weighted Gray-Edge~\cite{gijsenij2012improving}, using bright pixels~\cite{joze2012role}, using bright and dark colors~\cite{cheng2014illuminant}. The second group includes learning-based methods such as gamut mapping~(pixel, edge, and intersection based)~\cite{finlayson2006gamut}, using neural networks~\cite{cardei2002estimating}, using high-level visual information~\cite{van2007using}, natural image statistics~\cite{gijsenij2007color}, Bayesian learning~\cite{gehler2008bayesian}, spatio-spectral learning~(maximum likelihood estimate, and with gen. prior)~\cite{chakrabarti2012color}, simplifying the illumination solution space~\cite{banic2015color, banic2015using, banic2015acolor}, using color/edge moments~\cite{finlayson2013corrected}, using regression trees with simple features from color distribution statistics~\cite{cheng2015effective}, performing various kinds of spatial localizations~\cite{barron2015convolutional, barron2017fast}, using convolutional neural networks~\cite{bianco2015color, shi2016deep, hu2017fc4}.

Statistics-based illumination estimation methods are less accurate than the learning-based ones, but they are faster and simpler to implement in embedded systems, which is one of the reasons for their widespread usage~\cite{deng2011source}. Although in the relevant literature it often appears as if they require no training, this is not true because they have parameter values that need to be fine-tuned in order to give higher accuracy. In this paper it is first shown that in most papers on illumination estimation the accuracy of statistics-based methods was not obtained by means of the necessary cross-validation, but by using the whole benchmark datasets for both training and testing, which leads to an unfair comparison between the methods. After that the corrected results are given for the best known benchmark datasets by performing the same cross-validation framework as for other learning-based methods. Finally, the so called green stability assumption is proposed that can be used to fine-tune the values of the parameters of the statistics-based methods by using only non-calibrated images without known ground-truth illumination. The obtained accuracy is practically the same as when using calibrated training images, but the whole process is much faster and it can be directly applied in practice.

The paper is structured as follows: Section~\ref{sec:methods} briefly describes the best known statistics-based methods, Section~\ref{sec:revisited} shows that their accuracy data should be revisited, Section~\ref{sec:assumption} proposes the green stability assumption, Section~\ref{sec:results} presents the results, and finally, Section~\ref{sec:conclusions} concludes the paper.


\section{Best known statistics-based methods}
\label{sec:methods}

Some of the best known statistics-based illumination estimation methods are centered around the Gray-world assumption and its extensions. Under this assumption the average scene reflectance is achromatic~\cite{buchsbaum1980spatial} and $\mathbf{e}$ is therefore calculated as
\begin{equation}
\frac{\int\textbf{f}(x)dx}{\int dx}=k\textbf{e}
\label{eq:gw}
\end{equation}
where $k\in[0, 1]$ is the reflectance amount with $0$ meaning no reflectance and $1$ meaning total reflectance. By adding the Minkowski norm $p$ to Eq.~\eqref{eq:gw}, the Gray-world method is generalized into the Shades-of-Gray method~\cite{finlayson2004shades}:
\begin{equation}
\left(\frac{\int\left(\textbf{f}(x)\right)^p dx}{\int dx}\right)^\frac{1}{p}=k\textbf{e}.
\label{eq:sog}
\end{equation}
Having $p=1$ results in Gray-world, while $p\to\infty$ results in White-patch~\cite{land1977retinex, funt2010rehabilitation}. In~\cite{van2007edge} Eq.~\eqref{eq:sog} was extended to the general Gray-world by introducing local smoothing:
\begin{equation}
\left(\frac{\int\left(\textbf{f}^\sigma(x)\right)^p dx}{\int dx}\right)^\frac{1}{p}=k\textbf{e}
\label{eq:ggw}
\end{equation}
where $\textbf{f}^\sigma=\textbf{f}\ast\textbf{G}^\sigma$ and $\textbf{G}^\sigma$ is a Gaussian filter with standard deviation $\sigma$. Another significant extension is the Grey-edge assumption, under which the scene reflectance differences calculated with derivative order $n$ are achromatic~\cite{van2007edge} so that
\begin{equation}
\left(\int \left| \frac{\partial^n\left(\textbf{f}^\sigma(x)\right)^p}{\partial^n x} \right| dx \right)^\frac{1}{p}=k\textbf{e}.
\label{eq:ge}
\end{equation}

The described Shades-of-Gray, general Gray-world, and Gray-edge methods have parameters and the methods' accuracy depends on how the values of these parameters are tuned. Nevertheless, in the literature it often appears as if they require no training~\cite{gijsenij2011computational,cheng2014illuminant}, which is then said to be an advantage. It may be argued that the parameter values are in most cases the same, but this is easily disproved. In~\cite{cheng2014illuminant} for methods mentioned in this section the best fixed parameter values were given for ten different datasets. These values are similar for some datasets, but overall they span two orders of magnitude. With such high differences across different datasets in mind, it is obvious that the parameter values have to be learned.


\section{Validation revisited}
\label{sec:revisited}

\subsection{Angular error}
\label{sec:angular}

Before recalculating the accuracy of the methods from the previous section, some introduction to the used measures is needed. From various proposed illumination estimation accuracy measures~\cite{gijsenij2009perceptual,finlayson2014reproduction,banic2015perceptual}, the angular error is most commonly used. It represents the angle between the illumination estimation vector and the ground-truth illumination. All angular errors obtained for a given method on a chosen dataset are usually summarized by different statistics. Because of the non-symmetry of the angular error distribution, the most important of these statistics is the median angular error~\cite{hordley2004re}. Angular errors below $3^\circ$ are considered acceptable~\cite{finlayson2005colour, fredembach2008bright}.

The ground-truth illuminations of benchmark dataset images are obtained by reading off calibration objects put in the image scene, e.g. a gray ball or a color checker. When a method is tested, these objects are masked out to prevent possible bias.

\subsection{The need for cross-validation}
\label{sec:need}

When it comes to accuracy obtained on benchmark datasets, the ones available in~\cite{gijsenij2011computational}, at~\cite{gijsenij2017online}, and in~\cite{barron2015convolutional} are the most widely copied and referenced. If, for example, the results obtained for Shades-of-Gray on the GreyBall dataset~\cite{ciurea2003large} are checked~\cite{gijsenij2017online}, the reported mean and median angular errors of $6.1^\circ$ and $5.3^\circ$, respectively, are obtained by setting $p$ to $12$ on all $15$ folds. However, by performing cross-validation i.e. by looking for the best $p$ on $14$ training folds, applying this to the test fold, and repeating it all $15$ times clearly shows that $p$ differs for various training sets. Overall, the mean and median errors for combined results of all test folds are $7.8^\circ$ and $7.2^\circ$, respectively, which differs from the reported results. Similar differences can be shown for other methods from Section~\ref{sec:methods} as well. For the sake of fair comparison with other illumination estimation methods, these accuracies are properly recalculated and they are provided in Section~\ref{sec:results} together with other results.


\section{The proposed assumption}
\label{sec:assumption}

\subsection{Practical application}
\label{sec:practical}

The methods mentioned in Section~\ref{sec:methods} are some of the most widely used illumination estimation methods~\cite{deng2011source} and this means that their parameters should preferably be appropriately fine-tuned before putting them in production. The best way to do this is to use a benchmark dataset, but because of dependence of Eq.~\eqref{eq:e} on $\boldsymbol{\rho}(\lambda)$, a benchmark dataset would be required for each used camera sensor. Since putting the calibration objects into image scenes and later extracting the ground-truth illumination is time consuming, it would be better, if possible, to perform some kind of unsupervised learning on non-calibrated images without known ground-truth illumination. This would save time and be of practical value.

\subsection{Motivation}
\label{sec:motivation}

\begin{figure}[htb]
    \centering
    
	\includegraphics[width=\linewidth]{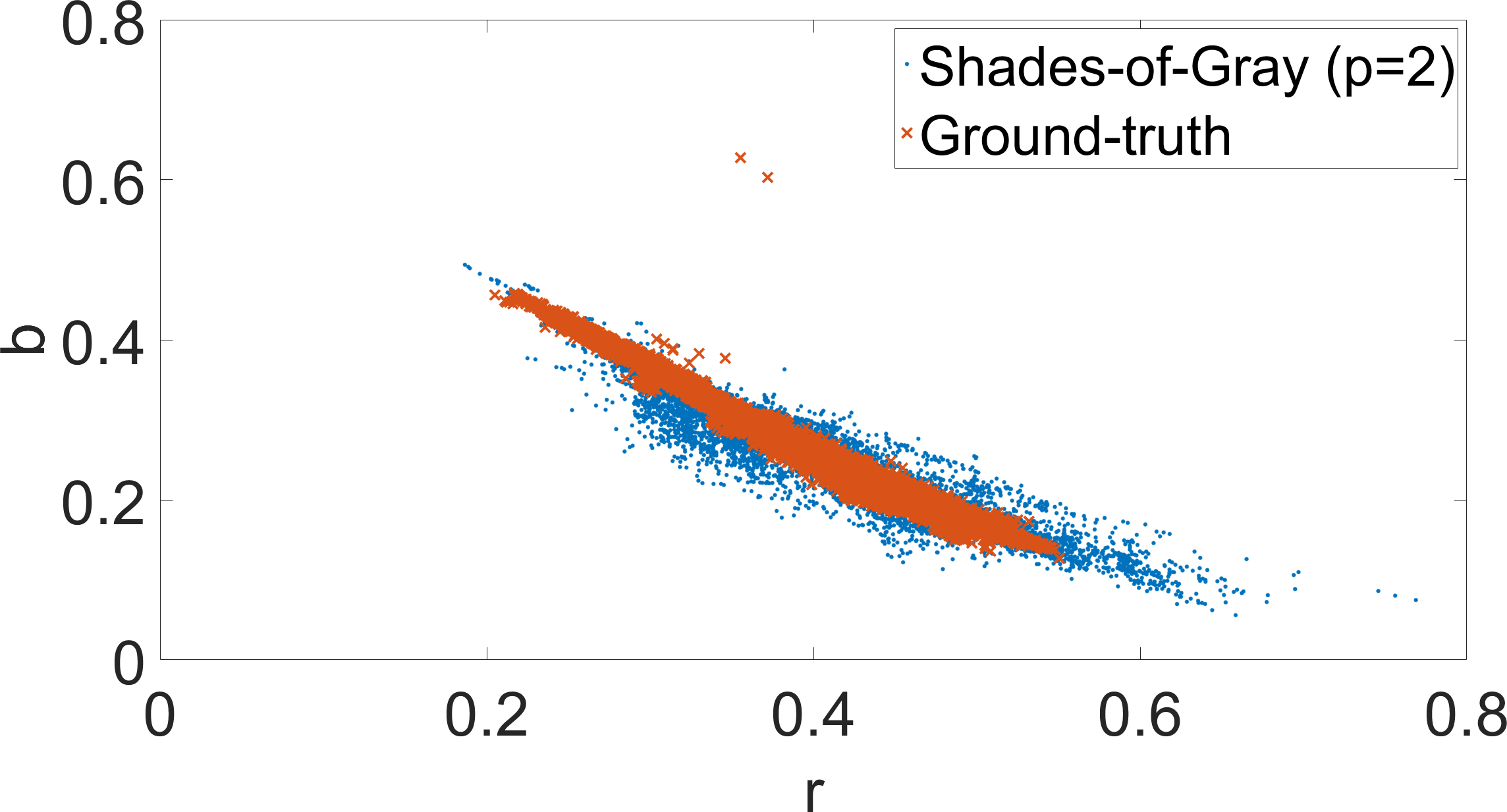}
	
    \caption{The $rb$-chromaticities of the ground-truth illuminations and Shades-of-Gray illumination estimations for GreyBall dataset images~\cite{ciurea2003large} (best viewed in color).}
	\label{fig:gt_sog}
    
\end{figure}

When for a dataset the ground-truth illuminations are unknown, an alternative is to make assumptions about the nature of illumination estimations produced by statistics-based methods when their parameters are fine-tuned and then to meet the conditions of the assumptions. When considering the nature of illumination estimations, a good starting point is the observation that some statistics-based illumination estimations appear ''to correlate \textit{roughly} with the actual illuminant''~\cite{finlayson2013corrected}. Fig.~\ref{fig:gt_sog} shows this for the images of the GreyBall dataset~\cite{ciurea2003large}.

\begin{figure}[htb]
    \centering
    
	\includegraphics[width=\linewidth]{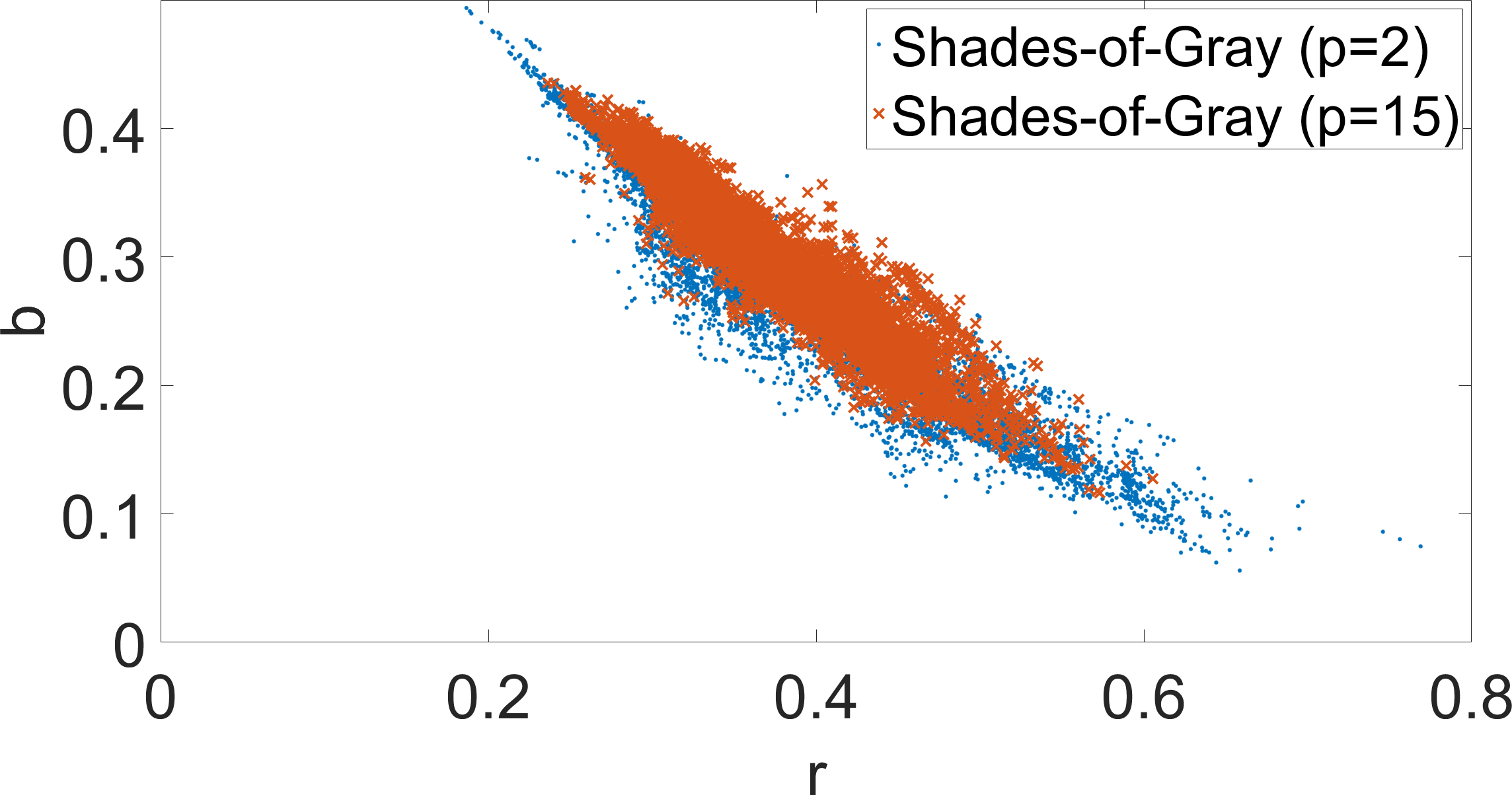}
	
    \caption{The $rb$-chromaticities of different Shades-of-Gray illumination estimations for GreyBall dataset images~\cite{ciurea2003large} (best viewed in color).}
	\label{fig:sogs}
    
\end{figure}

The points in Fig.~\ref{fig:gt_sog} can be considered to occupy a space around a line in the $rb$-chromaticity~\cite{banic2015color} which is connected to the fact that the green chromaticity of the ground-truth illuminations is relatively stable and similar for all illuminations. For the GreyBall dataset the standard deviations of the red, green, and blue chromaticity components of the ground-truth illuminations are $0.0723, 0.0106$, and $0.0750$, respectively, and similar results are obtained for all other datasets. For Shades-of-Gray illumination estimations shown in Fig.~\ref{fig:gt_sog} the red, green, and blue chromaticity components of the ground-truth illuminations are $0.0842, 0.0253$, and $0.0770$, respectively, which means that there is also a trend of green chromaticity stability, although the standard deviation is greater than in the case of ground-truth illumination. This means that if a set of illumination estimations is to resemble the set of ground-truth illuminations, the estimations' green chromaticity standard deviation should also be smaller and closer to the one of the ground-truth. As a matter of fact, if for example the Shades-of-Gray illumination estimations for $p=2$ and $p=15$ shown in Fig.~\ref{fig:sogs} are compared, the standard deviations of their green chromaticities are $0.0253$ and $0.0158$, respectively, while their median angular errors are $6.2^\circ$ and $5.3^\circ$, respectively. Similar behaviour where lower green chromaticity standard deviation is to some degree followed by lower median angular error can be seen on all datasets and for all methods from Section~\ref{sec:methods}.

\begin{figure}[htb]
    \centering
    
	\includegraphics[width=\linewidth]{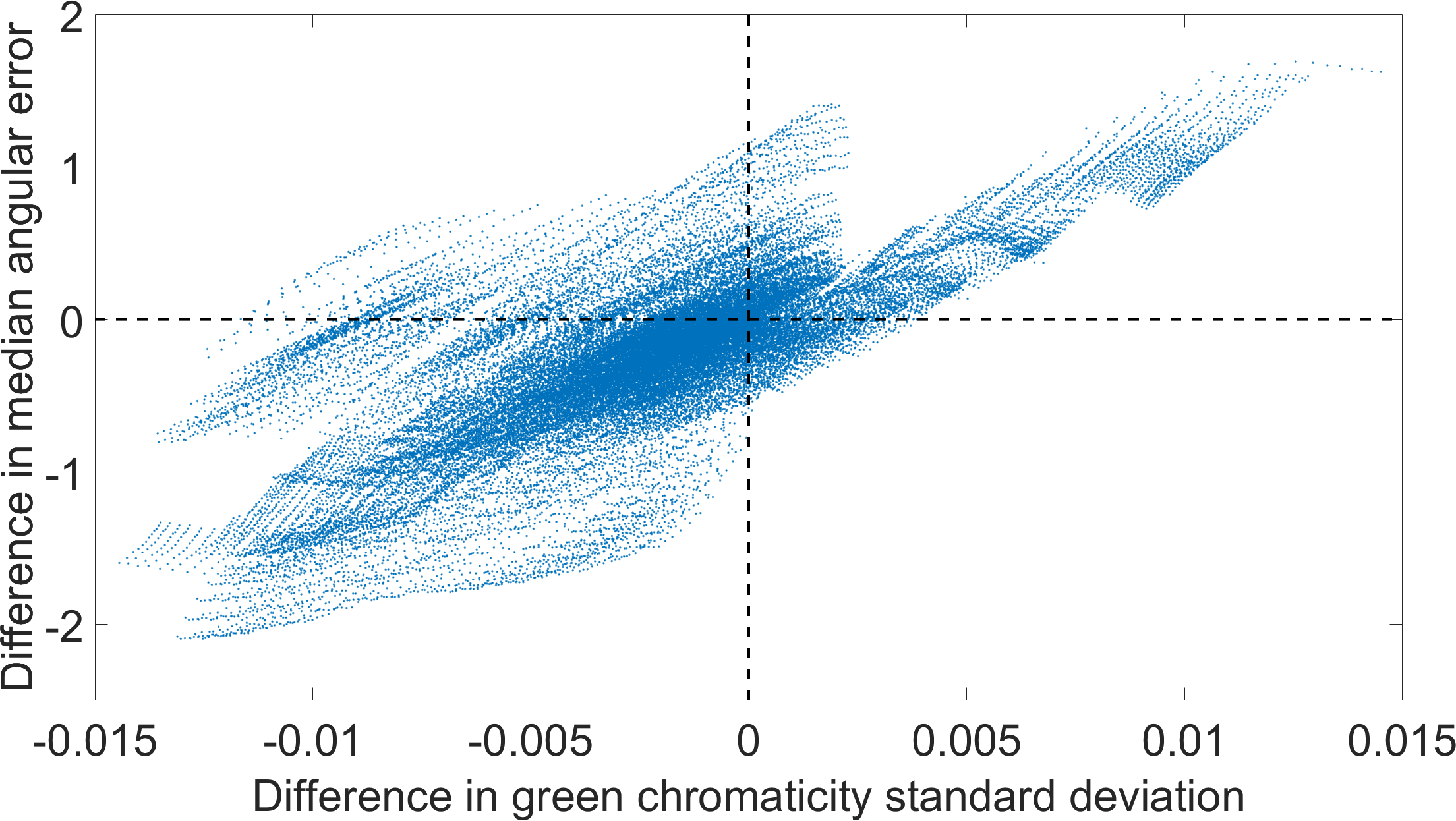}
	
    \caption{Relation between difference in standard deviations of illumination estimations' green chromaticity and the difference in illumination estimations' median angular error.}
	\label{fig:relation}
    
\end{figure}

For a deeper insight into this behaviour, another experiment was conducted on the GreyBall dataset~\cite{ciurea2003large}. First, for each method $M\in\mathbb{M}$, where $\mathbb{M}$ contains all methods from Section~\ref{sec:methods}, the Cartesian product of discrete sets of evenly spread values for individual parameters of $M$ was calculated to get $n$ tuples $\mathbf{p}_{M}^{(i)}, i\in \{1,2,\dots,n\}$. Gray-world and White-patch have no parameters, but they were implicitly included as special cases of Shades-of-Gray. Second, each $\mathbf{p}_{M}^{(i)}$ was used to set the parameter values of $M$ and then $M$ was applied to all images of the GreyBall dataset to obtain an illumination estimation for each of them. Third, for these illumination estimations the standard deviation of their green chromaticities $\sigma_i$ and their median angular error $m_i$ were calculated. Fourth, for every of ${n \choose 2}$ possible pairs of indices $i, j \in \{1,2,\dots,n\}$ such that $i<j$ a new difference pair $\{\Delta \sigma_k, \Delta m_k\}$ was calculated such that $\Delta \sigma_k=\sigma_i-\sigma_j$ and $\Delta m_k=m_i-m_j$. Finally, all such difference pairs created for all $M\in\mathbb{M}$ were put together into set of pairs $\mathbb{P}$. If members of pairs in $\mathbb{P}$ are interpreted as coordinates, then their plot is shown in Fig.~\ref{fig:relation}.

\subsection{Green stability assumption}
\label{sec:green}

The value of Person's linear correlation coefficient for the points in Fig.~\ref{fig:relation} is $0.7408$, which indicates a strong positive linear relationship~\cite{rumsey2015u}. In other words, the difference between the standard deviations of green chromaticities of illumination estimations produced by the same method when using different parameter values is strongly correlated to the difference between median angular errors of these illumination estimations. The same correlations for NUS datasets~\cite{cheng2014illuminant} are in Table~\ref{tab:correlation}.

\begin{table}[ht]
\tiny
\caption{Correlation between difference in green chromaticity standard deviation and difference in median angular errors for NUS datasets~\cite{cheng2014illuminant}.}
\label{tab:correlation}
\centering
\begin{tabular}{|c|c|c|c|c|c|c|c|c|}

	\hline
	\textbf{Dataset} & \textbf{C1} & \textbf{C2} & \textbf{Fuji} & \textbf{N52} & \textbf{Oly} & \textbf{Pan} & \textbf{Sam} & \textbf{Sony}\\
    \hline
	\textbf{Correlation} & 0.9255 & 0.6381 & 0.8977 & 0.9443 & 0.8897 & 0.9644 & 0.8902 & 0.9095\\
    \hline
    
\end{tabular}
\end{table}

Based on this empirical results and observations, it is possible to introduce the \textbf{green stability assumption}: the parameter values for which a method's illumination estimations' green chromaticity standard deviation is lower simultaneously lead to lower illumination estimation errors. Like many other assumptions, this assumption does not always hold, but it can still be useful in cases when the ground-truth illuminations for a set of images taken with a given sensor are not available. These images should also be taken under similar illuminations as the mentioned datasets that were used for empirical results.

For a specific case when the parameter values of a chosen method are fine-tuned and only non-calibrated images are available, the green stability assumption can be expressed more formally. If $n$ is the number of images in the training set, $\mathbf{p}_i$ is the $i$-th vector of parameter values, $\mathbf{e}^{i, j}$ is the method's illumination estimation obtained for the $j$-th image when $\mathbf{p}_i$ is used for parameter values, $e_G^{i, j}$ is the green component of $\mathbf{e}^{i, j}$, and $\overline{e_G^{i}}$ is the mean green component of illumination estimations for all images obtained with parameters $\mathbf{p}_i$, then under the green stability assumption the index $i^{*}$ of such $\mathbf{p}_{i^{*}}$ that should result in minimal angular errors is obtained as
\begin{equation}
\label{eq:i}
i^{*}=\underset{i}{\operatorname{\arg\min}} \sqrt{\frac{\sum_{j=1}^{n} \left( e_G^{i, j}-\overline{e_G^{i}} \right)^2}{n-1}}.
\end{equation}

Since Eq.~\eqref{eq:i} performs minimization of standard deviation, it can also be written without the square and the denominator.


\begin{table}[ht]
\scriptsize
\caption{Combined accuracy on eight NUS dataset (lower Avg. is better). The used format is the same as in~\cite{barron2017fast}.}
\label{tab:nus}
\centering
\begin{tabular}{|>{\centering}m{3.75cm}| >{\centering}m{3mm} >{\centering}m{3mm} >{\centering\arraybackslash}m{3mm} >{\centering\arraybackslash}m{3mm} >{\centering\arraybackslash}m{3mm} >{\centering\arraybackslash}m{3mm}|}

    \hline
    \textbf{Algorithm} & \textbf{Mean} & \textbf{Med.} & \textbf{Tri.} & \makecell{\textbf{Best}\\\textbf{25\%}} & \makecell{\textbf{Worst}\\\textbf{25\%}} & \textbf{Avg.}\\
    
    \hline
	\multicolumn{7}{|c|}{\textbf{Originally reported results}}\\    
	\hline
    Shades-of-Gray~\cite{finlayson2004shades} & 3.67 & 2.94 & 3.03 & 0.98 & 7.75 & 3.01\\
    General Gray-World~\cite{barnard2002comparison} & 3.20 & 2.56 & 2.68 & 0.85 & 6.68 & 2.63\\
    1st-order Gray-Edge~\cite{van2007edge} & 3.35 & 2.58 & 2.76 & 0.79 & 7.18 & 2.67\\
    2nd-order Gray-Edge~\cite{van2007edge} & 3.36 & 2.70 & 2.80 & 0.89 & 7.14 & 2.76\\
    \hline
	\multicolumn{7}{|c|}{\textbf{Revisited results}}\\    
	\hline
    Shades-of-Gray~\cite{finlayson2004shades} & 3.48 & 2.63 & 2.81 & 0.81 & 7.62 & 2.76`\\
    General Gray-World~\cite{barnard2002comparison} & 3.37 & 2.49 & 2.61 & 0.73 & 7.58 & 2.61\\
    1st-order Gray-Edge~\cite{van2007edge} & 3.12 & 2.19 & 2.39 & 0.71 & 7.11 & 2.42\\
    2nd-order Gray-Edge~\cite{van2007edge} & 3.15 & 2.23 & 2.42 & 0.74 & 7.13 & 2.46\\
    \hline
	\multicolumn{7}{|c|}{\textbf{Green stability assumption results}}\\    
	\hline
    Shades-of-Gray~\cite{finlayson2004shades} & 3.44 & 2.65 & 2.81 & 0.83 & 7.41 & 2.75\\
    General Gray-World~\cite{barnard2002comparison} & 3.40 & 2.63 & 2.76 & 0.77 & 7.42 & 2.69\\
    1st-order Gray-Edge~\cite{van2007edge} & 3.29 & 2.36 & 2.55 & 0.79 & 7.36 & 2.58\\
    2nd-order Gray-Edge~\cite{van2007edge} & 3.29 & 2.44 & 2.59 & 0.83 & 7.30 & 2.63\\
    \hline
    
\end{tabular}
\end{table}

\begin{table}[ht]
\scriptsize
\caption{Accuracy on the original GreyBall dataset (lower median is better).}
\label{tab:gb}
\centering
\begin{tabular}{|c|c|c|c|}
    \hline
    \textbf{method} & \textbf{mean $(^{\circ})$} & \textbf{median $(^{\circ})$} & \textbf{trimean $(^{\circ})$}\\
    
    \hline
    \hline
	\multicolumn{4}{|c|}{\textbf{Originally reported results}}\\
	\hline
    Shades-of-Gray~\cite{finlayson2004shades} & 6.14 & 5.33 & 5.51\\
    \hline
    General Gray-World~\cite{barnard2002comparison} & 6.14 & 5.33 & 5.51\\
    \hline
    1st-order Gray-Edge~\cite{van2007edge} & 5.88 & 4.65 & 5.11\\
    \hline
    2nd-order Gray-Edge~\cite{van2007edge} & 6.10 & 4.85 & 5.28\\
    \hline
	\multicolumn{4}{|c|}{\textbf{Revisited results}}\\
    \hline
    Shades-of-Gray~\cite{finlayson2004shades} & 7.80 & 7.15 & 7.21\\
    \hline
    General Gray-World~\cite{barnard2002comparison} & 7.61 & 6.85 & 6.92\\
    \hline
    1st-order Gray-Edge~\cite{van2007edge} & 6.14 & 5.32 & 5.49\\
    \hline
    2nd-order Gray-Edge~\cite{van2007edge} & 6.89 & 5.84 & 6.06\\
    \hline
	\multicolumn{4}{|c|}{\textbf{Green stability assumption results}}\\
	\hline
    Shades-of-Gray~\cite{finlayson2004shades} & 6.80 & 5.30 & 5.77\\
    \hline
    General Gray-World~\cite{barnard2002comparison} & 6.80 & 5.30 & 5.77\\
    \hline
    1st-order Gray-Edge~\cite{van2007edge} & 5.97 & 4.64 & 5.10\\
    \hline
    2nd-order Gray-Edge~\cite{van2007edge} & 6.69 & 5.17 & 5.72\\
	\hline
	
\end{tabular}
\end{table}

\begin{table}[ht]
\scriptsize	
\caption{Accuracy on the linear GreyBall dataset (lower median is better).}
\label{tab:lgb}
\centering
\begin{tabular}{|c|c|c|c|}
    \hline
    \textbf{method} & \textbf{mean $(^{\circ})$} & \textbf{median $(^{\circ})$} & \textbf{trimean $(^{\circ})$}\\
    
    \hline
    \hline
	\multicolumn{4}{|c|}{\textbf{Originally reported results}}\\
	\hline
    Shades-of-Gray~\cite{finlayson2004shades} & 11.55 & 9.70 & 10.23\\
    \hline
    General Gray-World~\cite{barnard2002comparison} & 11.55 & 9.70 & 10.23\\
    \hline
    1st-order Gray-Edge~\cite{van2007edge} & 10.58 & 8.84 & 9.18\\
    \hline
    2nd-order Gray-Edge~\cite{van2007edge} & 10.68 & 9.02 & 9.40\\
    \hline
	\multicolumn{4}{|c|}{\textbf{Revisited results}}\\
    \hline
    Shades-of-Gray~\cite{finlayson2004shades} & 13.32 & 11.57 & 12.10\\
    \hline
    General Gray-World~\cite{barnard2002comparison} & 13.69 & 12.11 & 12.55\\
    \hline
    1st-order Gray-Edge~\cite{van2007edge} & 11.06 & 9.54 & 9.81\\
    \hline
    2nd-order Gray-Edge~\cite{van2007edge} & 10.73 & 9.21 & 9.49\\
    \hline
	\multicolumn{4}{|c|}{\textbf{Green stability assumption results}}\\
	\hline
    Shades-of-Gray~\cite{finlayson2004shades} & 12.68 & 10.50 & 11.25\\
    \hline
    General Gray-World~\cite{barnard2002comparison} & 12.68 & 10.50 & 11.25\\
    \hline
    1st-order Gray-Edge~\cite{van2007edge} & 13.41 & 11.04 & 11.87\\
    \hline
    2nd-order Gray-Edge~\cite{van2007edge} & 12.83 & 10.70 & 11.44\\
	\hline
    
\end{tabular}
\end{table}

\section{Experimental results}
\label{sec:results}

\subsection{Experimental setup}
\label{subsec:setup}

The following benchmark datasets have been used to demonstrate the difference between previously reported and newly calculated accuracy results for methods mentioned in Section~\ref{sec:methods} and to test the effectiveness of the proposed green stability assumption: the GreyBall dataset~\cite{ciurea2003large}, its approximated linear version, and eight linear NUS dataset~\cite{cheng2014illuminant}. The ColorChecker dataset~\cite{gehler2008bayesian, shi2015online} was not used because of its confusing history of wrong usage despite warnings from leading experts~\cite{lynch2013colour}. Except the original GreyBall dataset, all other contain linear images, which is preferred because illumination estimation is in cameras usually performed on linear images~\cite{kim2012new} similar to the model described by Eq.~\eqref{eq:image}.

The tested methods include all the ones from $\mathbb{M}$. During cross-validation on all datasets the same folds were used as in other publications. The source code for recreating the numerical results given in the following subsection is publicly available at \url{http://www.fer.unizg.hr/ipg/resources/color_constancy/}.

\subsection{Accuracy}
\label{subsec:accuracy}

Tables~\ref{tab:nus},~\ref{tab:gb}, and~\ref{tab:lgb} show the previously reported accuracies, the newly recalculated accuracies, and the accuracies obtained by using the green stability assumption. The results clearly confirm the potential and the practical applicability of the green stability assumption. This also demonstrates the success of unsupervised learning for illumination estimation.


\section{Conclusions and future research}
\label{sec:conclusions}

In most relevant papers the accuracy results for some of the most widely used statistics-based methods were calculated without cross-validation. Here it was shown that cross-validation is needed and the accuracy results were revisited. When statistics-based methods are fine-tuned, the best way to do this is by using images with known ground-truth illumination. Based on several observations and empirical evidence, the green stability assumption has been proposed that can be successfully used to fine-tune the parameters of statistics-based methods when only non-calibrated images without ground-truth illumination are available. This makes the whole fine-tuning process much simpler, faster, and more practical. It is also an unsupervised learning approach to color constancy. In future, other similar bases for further assumptions for unsupervised learning for color constancy will be researched.

\balance
\bibliographystyle{IEEEtran}
\bibliography{g}

\end{document}